\documentclass[journal]{IEEEtran}

\usepackage{amssymb}
\usepackage{mathrsfs}
\usepackage{bbm}
\usepackage{subfigure}
\usepackage{graphicx}
\graphicspath{{figures/}}
\usepackage{picins}
\usepackage{multirow}
\usepackage{threeparttable}
\usepackage{makecell}
\usepackage{booktabs}
\usepackage{amsmath}
\usepackage{float}
\usepackage[T1]{fontenc}

\usepackage[linesnumbered, ruled,vlined]{algorithm2e}
\usepackage{algorithmic}

\newtheorem{proof}{Proof}
\usepackage{colortbl}
\usepackage{xcolor}
\usepackage{color}
\usepackage[colorlinks,linkcolor=red,anchorcolor=blue,citecolor=green]{hyperref}

\allowdisplaybreaks[4]

\usepackage{tikz}
\definecolor{lime}{HTML}{A6CE39}
\DeclareRobustCommand{\orcidicon}{%
	\begin{tikzpicture}
		\draw[lime, fill=lime] (0,0) 
		circle [radius=0.16] 
		node[white] {{\fontfamily{qag}\selectfont \tiny ID}};    \draw[white, fill=white] (-0.0625,0.095) 
		circle [radius=0.007];    \end{tikzpicture}
	\hspace{-2mm}}
\foreach \x in {A, ..., Z}{%
	\expandafter\xdef\csname orcid\x\endcsname{\noexpand\href{https://orcid.org/\csname orcidauthor\x\endcsname}{\noexpand\orcidicon}}
}

\begin{document}

\title{Differentiated Information Mining: A Semi-supervised Learning Framework for GNNs}
\author{Long~Wang\orcidA{}, 
        Kai~Liu\orcidB{},
}

\maketitle

\begin{abstract}

In semi-supervised learning (SSL) for enhancing the performance of graph neural networks (GNNs) with unlabeled data, introducing mutually independent decision factors for cross-validation is regarded as an effective strategy to alleviate pseudo-label confirmation bias and training collapse. However, obtaining such factors is challenging in practice: additional and valid information sources are inherently scarce, and even when such sources are available, their independence from the original source cannot be guaranteed. To address this challenge, In this paper we propose a Differentiated Factor Consistency Semi-supervised Framework (DiFac), which derives differentiated factors from a single information source and enforces their consistency. During pre-training, the model learns to extract these factors; in training, it iteratively removes samples with conflicting factors and ranks pseudo-labels based on the “shortest stave” principle, selecting the top candidate samples to reduce overconfidence commonly observed in confidence-based or ensemble-based methods. Our framework can also incorporate additional information sources. In this work, we leverage the large multimodal language model to introduce latent textual knowledge as auxiliary decision factors, and we design a accountability scoring mechanism to mitigate additional erroneous judgments introduced by these auxiliary factors. Experiments on multiple benchmark datasets demonstrate that DiFac consistently improves robustness and generalization in low-label regimes, outperforming other baseline methods.

\end{abstract}


\IEEEpeerreviewmaketitle

\section{Introduction}
\label{Sec:Introduction}

Graph Neural Networks (GNNs) have demonstrated remarkable performance across various graph-structured data analysis tasks, such as node classification, link prediction, and graph classification, by effectively integrating node attributes with structural information~\cite{wu2020comprehensive}. However, their success largely depends on the availability of large-scale, high-quality labeled datasets, which are often expensive or even infeasible to obtain in real-world scenarios. Semi-Supervised Learning (SSL) offers an effective remedy by leveraging abundant unlabeled data to improve model generalization, especially in low-label settings~\cite{wang2021confident,luo2023toward}.

Among existing SSL techniques, pseudo-labeling has been widely adopted for its simplicity and scalability~\cite{li2023informative}. This approach assigns “pseudo-labels” to unlabeled samples based on high-confidence predictions and incorporates them into the training set to strengthen supervision. However, pseudo-labeling is prone to confirmation bias and training collapse~\cite{arazo2020pseudo,cascante2021curriculum}: incorrect early predictions, once accepted, are recursively reinforced in subsequent training. In GNNs, the message-passing mechanism amplifies this issue by propagating local errors across the network. Moreover, the confidence estimates of deep models are not always reliable—high confidence does not guarantee correctness—making overconfidence particularly harmful in graph models~\cite{li2018deeper}.

Existing solutions can be broadly categorized into ensemble-based methods and consistency regularization methods. Ensemble-based methods train multiple models or apply multiple perturbations to the same model, and select pseudo-labels based on consensus or voting~\cite{zhang2021flexmatch}. While such approaches can alleviate overconfidence, they incur substantial computational costs. Consistency regularization methods~\cite{rizvedefense,li2023informative} enhance prediction stability under perturbations to improve robustness, but still rely on potentially flawed confidence scores. Both paradigms implicitly assume access to multiple independent “views” or factors, which is inherently difficult to achieve in many real-world scenarios—especially in single-source environments.

To address this challenge, we propose a heuristic sample selection framework based on differentiated factor consistency from a single information source (Differentiating-Factor Consistency Semi-
supervised Framework, DiFac). During pretraining, the model learns multiple independent or weakly correlated factors from the same source, each serving as an “internal view” of the data. In pseudo-label generation, we first perform consistency filtering: samples with conflicting predictions across factors are discarded, retaining only those with high cross-factor agreement. This selective acceptance mitigates noisy supervision, alleviates confirmation bias, and preserves the integrity of learned representations—without requiring additional models.

Building upon this, we introduce a Pseudo-Label Ranking mechanism to further refine sample selection after consistency filtering. Unlike traditional approaches that rank by the highest confidence score, we use the minimum confidence across all factors as the ranking criterion. This means a sample’s final rank is determined by its weakest supporting factor, ensuring that all “views” provide sufficiently strong evidence for the label. This strategy not only theoretically suppresses the negative effects of overconfidence from individual factors but also yields more stable performance improvements in practice.

Furthermore, our framework naturally extends to multi-source integration. Additional information such as graph topology, node textual descriptions, visual features, or external knowledge bases—can serve as extra factors to enrich decision-making. Recent advances in multimodal large models make it possible to generate augmented descriptions of nodes or edges, providing additional factors to enhance sample selection robustness. However, the predictive quality of additional sources can vary widely, and directly involving them in pseudo-labeling may degrade overall performance due to noisy or misleading information. To address this, we propose an Accountability Scoring mechanism: external factors do not participate in the initial filtering stage but contribute their confidence scores only during the ranking process. This “limited involvement” preserves the potential benefits of auxiliary information while avoiding its direct interference with initial selection, striking a balance between leveraging multi-source information and maintaining decision stability.

Extensive experiments on benchmark citation networks such as Cora, Citeseer, and Pubmed, as well as other graph datasets, demonstrate that our method consistently improves GNN robustness and generalization in low-label regimes, outperforming conventional pseudo-labeling and confidence-based SSL approaches while avoiding the computational overhead of ensemble training.

The contributions of our work are as follows:

\begin{itemize}

\item We propose a novel pseudo-label sample selection framework based on the consistency of differentiated factors derived from a single information source, effectively mitigating confirmation bias and learning collapse inherent in traditional pseudo-labeling methods.

\item We design an iterative selective acceptance mechanism that ranks candidate samples by the minimum confidence across multiple factors, prioritizing reliable samples for subsequent training iterations. This strategy effectively suppresses the adverse effects of model overconfidence on pseudo-label reliability.

\item We demonstrate the efficacy of augmented data from large models as independent decision factors, paving the way for future extensions incorporating multimodal information sources. Additionally, we introduce an accountability scoring mechanism that preserves the potential benefits of auxiliary information while preventing its direct interference during initial pseudo-label selection.

\item To comprehensively evaluate our semi-supervised learning framework, we conduct extensive experiments on multiple benchmark graph datasets. Results show that under low-label and high-noise conditions, our method consistently outperforms existing pseudo-labeling and semi-supervised approaches, achieving superior accuracy, enhanced robustness, and improved training stability, thereby demonstrating strong practical value and application potential.

\end{itemize}

This work not only offers a novel perspective for semi-supervised graph learning research but also contributes both theoretical insights and practical value toward developing more reliable machine learning systems.

The remainder of the paper is organized as follows: Section~\ref{Sec:RelatedWork} reviews related work. Section~\ref{sec:Method} presents the Differentiated Factor Consistency Sample Selection Framework (DiFac) and its application on auxiliary information sources from LMM. Efficiency comparisons between DiFac and manifold-based baseline models are provided in Section~\ref{sec:Experiment}. Finally, conclusions and future directions are discussed in Section~\ref{sec:Conclusion}.

\section{Related Works}
\label{Sec:RelatedWork}

In this section, we review recent advances in graph semi-supervised learning and knowledge enhancement via large multimodal models (LMMs), broadly divided into the following subsections: Graph-oriented Gemi-supervised Learning Techniques, and Data Enrichment via LMMs.

\subsection{graph-oriented semi-supervised learning techniques}
\label{SubSec:Multi_Cooperation}
The effectiveness of Graph Neural Networks (GNNs) has been demonstrated across a wide range of tasks, including node classification, link prediction, and graph-level representation learning. Despite their success, GNNs often require substantial labeled data, which is scarce in many practical applications, thereby motivating research into semi-supervised approaches that leverage unlabeled data efficiently.
Pseudo-label self-training has become a popular approach to address label scarcity in semi-supervised node classification with Graph Neural Networks (GNNs). Yu et al.\cite{YuSDZL23} enhance this by treating labels as virtual centers for joint node-label representation learning and adopt adaptive self-training to iteratively expand the pseudo-label set, improving class separability and generalization. To mitigate pseudo-label noise, Lu et al.\cite{LuGZYLXYT25} propose Pseudo Contrastive Learning (PCL), converting classification supervision into negative contrastive constraints between topologically close but differently predicted nodes, along with a topology-weighted loss to improve robustness across GNNs. Graph fusion models use pseudo-label supervision for joint learning on heterogeneous graphs to boost cross-graph transfer~\cite{2024Improving}. Combining self-training with active learning, SEG introduces an informative pseudo-label sampling mechanism~\cite{YangSJWGY23}. Additionally, Wang et al. proposes adaptive, class-wise confidence thresholds to filter pseudo-labels, addressing class imbalance and enhancing unlabeled data use~\cite{YuWLWW23}.

\subsection{Data Enrichment via LMMs} 
\label{SubSec:Data_Augmentation}

Recently, large language models (LLMs) have demonstrated remarkable capabilities in data synthesis and augmentation, particularly in scenarios with limited labeled data~\cite{min2023recent, wang2025ttvae}. These models leverage their vast pre-trained knowledge to generate high-quality synthetic data that can improve downstream learning tasks. For instance, Wang et al.\cite{qu2020coda} proposed the CoDA framework, which integrates contrastive learning with diversity-promoting strategies to produce more informative and diverse training datasets, thereby enhancing natural language understanding. In the vision domain, LLaS++ utilizes extensive prior knowledge to generate pseudo-labels and incorporates curriculum learning techniques to progressively refine data quality, resulting in improved image segmentation outcomes~\cite{zhu2025llafs++}. In healthcare applications, Med-PaLM achieves expert-level performance in medical question answering by embedding rich clinical information within large-scale language representations~\cite{singhal2023large}. Furthermore, Frank et al.\cite{frank2024leveraging} combined embeddings derived from LLMs with physical and multiscale modeling to accurately predict complex protein phase separation phenomena. Collectively, these advances highlight the broad and versatile potential of LLMs for enhancing data generation and augmentation across diverse fields, driving improved performance in both natural language and biomedical domains~\cite{zang2025contextual, duval2024phast}.

\section{Methodology}
\label{sec:Method}
In this section, we introduce our proposed unlabeled data learning framework DiFac and discuss the advantages of our framework design.

\begin{figure*}
  \centering
    \includegraphics[width=\textwidth]{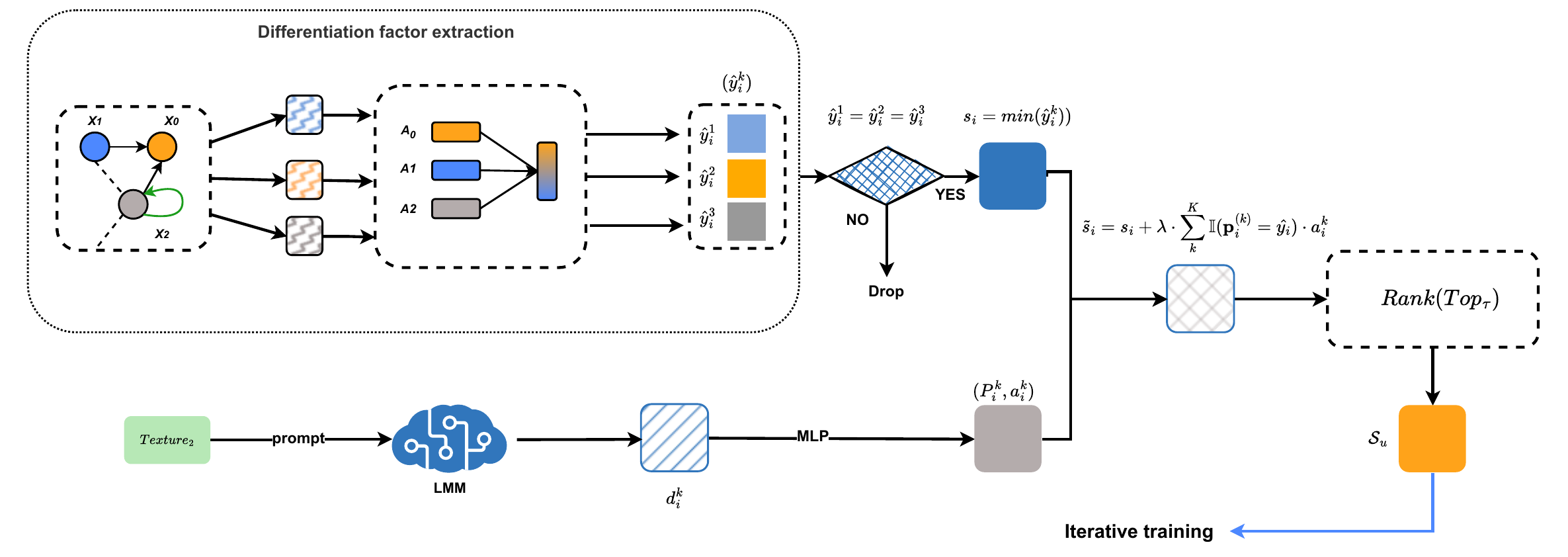}
  \caption{CAnNN Framework. 
  Let $\{x_0,x_1,x_2\}$ denote structured inputs, which, through the aforementioned guidance process, yield multiple differentiated factor predictions $\hat{y}^k_i$. The score $s_i$ is computed as the minimum value among these outputs. Auxiliary judgments and scores $(P_i^k,a_i^k)$ are provided by large multimodal models (LMMs) based on additional information such as textual descriptions. The combined score $\tilde{s}_i$ is produced via the accountability scoring mechanism.
  }
  \label{fig:framework}
\end{figure*}

We first partition the single-source information into distinct input categories by incorporating labels, thereby guiding the neural network to extract diverse, differentiated factors. These factors, obtained during pretraining, are subsequently employed for learning on unlabeled data, where we introduce a ranking strategy to selectively identify reliable samples. Furthermore, auxiliary information generated by large multimodal models (LMMs) is integrated through an accountability scoring mechanism during unlabeled learning, enhancing the robustness and reliability of overall predictions.
The overall framework of our proposed DiFac is shown in Fig \ref{fig:framework}.

\subsection{Differentiated Information Mining}

The Condorcet Jury Theorem states that if multiple classifiers each achieve an accuracy slightly higher than random guessing (e.g., > 50$\%)$ and their decisions are mutually independent, then aggregating their outputs via majority voting will cause the overall accuracy to rapidly approach 100$\%)$ as the number of judges increases.

\label{subsec:C_C_Extraction}
\begin{proof}

Consider a binary classification problem for variable $A$ with classes $\{0,1\}$,We define two events accordingly:

\begin{equation}
\begin{aligned}
     H_1 :Class_A  = 1
     \\
    H_0 :Class_A = 0
    \\
\end{aligned}
\end{equation}

\begin{equation}
     P(H_1) = \mathrm{\pi},P(H_0) =1- \mathrm{\pi}
\end{equation}

The decision model achieves a correct judgment with probability $p_d$, and an incorrect judgment with probability $(1-p_d)$. The exclusion model correctly allows passage with probability $p_e$, while it erroneously allows passage with probability $(1-p_e)$, meaning it fails to exclude an option that should have been filtered out.

Our goal is that when the sample belongs to the positive category $H_1$, the model correctly judges $(D)$ and the exclusion model also releases $(E)$, $P(H_1 \mid D, E)$, which is expanded by Bayes' theorem:

\begin{equation}
    P(H_1 \mid D, E) = \frac{P(D, E \mid H_1) \cdot P(H_1)}{P(D, E)}
\end{equation}
\begin{equation}
    P(D, E) = P(D, E \mid H_1) \cdot P(H_1) + P(D, E \mid H_0) \cdot P(H_0)
\end{equation}

When events $D$ and $E$ are independent of each other in $H$, we have:

\begin{equation}
\begin{aligned}
      P(D|H_1)=P_d, P(E|H_1)=p_e 
      \\
      P(D,E|H_1)=p_d \cdot p_e
      \\
      P(D|H_0) = 1-p_d,P(E|H_0) = 1-p_e
      \\
      P(D,E|H_0) = (1-p_d)(1-p_e)
\end{aligned}
\end{equation}

Substitute into the Bayesian formula to get the posterior:

\begin{equation}
    P(H_1 \mid D, E) = \frac{p_d \cdot p_e \cdot \pi}{p_d \cdot p_e \cdot \pi + (1 - p_d) \cdot (1 - p_e) \cdot (1 - \pi)}
\end{equation}

Similarly, we can get the posterior of only the judgment model:

\begin{equation}
    P(H_1 \mid D) = \frac{p_d \cdot \pi}{p_d \cdot \pi + (1 - p_d) \cdot (1 - \pi)}
\end{equation}

Solve the conditions for the exclusion model to take effect:

\begin{equation}
    P(H_1 \mid D, E) - P(H_1 \mid D) > 0
\end{equation}

Substituting in and sorting out, we can get:

\begin{equation}
\begin{aligned}
    p_d \pi(1-P_d)(1-\pi)(2p_e-1)>0
    \\
    2p_e-1>0
    \\
    p_e>0.5
\end{aligned}
\end{equation}

It is concluded that as long as the exclusion judgment of the exclusion model is better than random $p_e>0.5$, the reliability of the existing judgment can be improved.
\end{proof}

Based on the conclusions derived above, it is evident that the more accurate the exclusion model’s decisions and the greater its independence from the decision model, the more it can enhance the reliability of existing judgments. To obtain independent factors for exclusion information, we propose a novel data augmentation technique and model architecture adjustment designed to fully exploit the information embedded within labeled data.

Factor identification information is added to the sample attribute $A \in \hat{X}$ to form a new attribute $\hat{A} \in \hat{X}$. Assuming one direct judgment factor and two exclusion judgment factors, we have $\hat{X}^k \in \{\hat{X}^0,\hat{X}^1,\hat{X}^2\}$. Correspondingly, the sample label $Y$ is expanded to $Y^k \in \{Y^0,Y^1,Y^2\}$. Leveraging the backpropagation algorithm's natural ability to distinguish between samples of different categories, each factor learned by the model is separated. The algorithm is showed as \ref{alg:TIF}.

\begin{algorithm}
\caption{Training of Independent Factors}
\label{alg:TIF}
\SetAlgoLined

\KwIn{Extended Data $\hat{X}^k$, Extended Labels $Y^k$, Number of differentiated factors $K$, Neural Parameter $\theta$, Depth $s$, Number of Training Iteration $N_{t-iter}$, Learning Rate $\eta$;}
\KwOut{Differentiated factors $f_k$:}
Initialize GCN$=p(\hat{X}|\theta)$\;
$H^{(0)k} \leftarrow (X^k) , \ k \in K$\;
  \For{$j=1$ to $N_{t-iter}$}
    {
        \For{$l=1$ to $s$}
        {$H^{(l)k} \leftarrow \sigma(H^{(l-1)k} \cdot \theta_l)$\;
        }
        $\hat{Y}^k \leftarrow  H^{(s)k}$\;
        $\mathcal{L}_{\theta} \leftarrow \frac{1}{|X|} \|Y^{k} \circ \log \hat{Y}^k \|_F $\;

        $\theta\leftarrow \theta - \eta \nabla_{\theta} \mathcal{L}_{\theta}$\;
    
    }
    $f_k : GCN(X^k) \rightarrow Y^k$\;
\Return $f_k$.
\end{algorithm}

\subsection{Pseudo Label Consistency}
\label{subsec:S_S_Repartitioning}

We propose a semi-supervised framework termed Differentiating-Factor Consistency Semi-supervised Framework (DiFac). The core idea is to leverage the prediction consistency among multiple, ideally independent, discriminative factors on unlabeled samples as a measure of reliability. By selecting high-confidence pseudo-labeled samples based on this consistency, DiFac effectively enhances the overall generalization capability of the model.

Assume that there are $K$ independent differentiated factors, which are recorded as:

\begin{equation}
    f_k: \mathcal{X} \rightarrow \mathcal{Y}, \quad k=1,2,\dots,K
\end{equation}

$f_k(x)$ represents the category prediction probability distribution of the $k$th factor for sample $x$.

In each iteration, the model makes predictions on the unlabeled data:

\begin{equation}
    \hat{y}_{j}^{(k)} =f_k(x_j)
\end{equation}

If all $K$ factors predict the same on sample $x_j$, that is:

\begin{equation}
    \hat{y}_{j}^{(1)} = \hat{y}_{j}^{(2)} = \dots = \hat{y}_{j}^{(K)} = \hat{y}_j
\end{equation}

Then add the samples and thier pseudo labels $\hat{y}_j$ to the next round of training set:

\begin{equation}
    \mathcal{D}^{(t+1)} = \mathcal{D}_{l} \cup \{(x_j, \hat{y}_j)\}
\end{equation}

Otherwise, the sample is discarded, and this process continues until model convergence. To formalize this consistency-based selection mechanism, we define a consistency indicator function:

\begin{equation}
    \mathbb{I}_j = 
\begin{cases}
1, & \text{if } \ \hat{y}_{j}^{(1)} = \hat{y}_{j}^{(2)} = \dots = \hat{y}_{j}^{(K)}\\[3pt]
0, & \text{otherwise}
\end{cases}
\end{equation}

The set of pseudo-labeled samples adopted in one round of iteration can be expressed as:

\begin{equation}
    \mathcal{S}_u = \{ (x_j, \hat{y}_j) \mid \mathbb{I}_j = 1,\, x_j \in \mathcal{D}_u \}
\end{equation}

Where $\mathcal{D}_u$ represents the unlabeled dataset.

The loss function finally used to update the model consists of two parts.

Supervised loss for labeled samples:

\begin{equation}
    \mathcal{L}_s = - \frac{1}{|\mathcal{D}_l|} \sum_{(x_i,y_i)\in\mathcal{D}_l} \sum_{c=1}^{C} {(y_i)_c}\log f_k(x_i)_c
\end{equation}

Pseudo-label loss for unlabeled consistent samples:

\begin{equation}
    \mathcal{L}_u = - \frac{1}{|\mathcal{S}_u|} \sum_{(x_j,\hat{y}_j)\in\mathcal{S}_u} \sum_{c=1}^{C} (\hat{y_j})_c\log f_k(x_j)_c
\end{equation}

The total joint loss is:

\begin{equation}
    \mathcal{L} = \mathcal{L}_s + \lambda \mathcal{L}_u
\end{equation}

Where $\lambda$ is the balance coefficient and $c\in C$ is the classification category.

The process of learning unlabeled data is as shown in algorithm \ref{alg:MFCSL}.

\begin{algorithm}
\caption{MFCSL with UnLabeled Data}
\label{alg:MFCSL}
\SetAlgoLined
\KwIn{Labeled data $D_l$, unlabeled data $D_u$}
\KwOut{$f_k$}
Initialize $K$ factor models {$f_k$} using $D_l$\;

\Repeat{convergence}{

    Su = \{\}\;
    \For{$x$ in $D_u$}
    {
     $\hat{y}^{(k)} =f_k(x), \ \ k \in \{1..K\}$\;
     $\hat{y}^{(k)} \rightarrow \mathbb{I}$\; 
     \If{$\mathbb{I} == 1$}{ $Su.add((x, \hat{y}))$}\;
    }
Update models using $D_l \cup S_u$\;
}
\Return $f_k$.
\end{algorithm}

\subsection{Rank pseudo-labeling strategy}
\label{subsec:c_p_prediction}

It is foreseeable that the model’s predictions on unlabeled data inevitably contain errors. Once erroneous pseudo-labels are generated during the early training iterations, they are used to update model parameters, reinforcing incorrect patterns and potentially causing learning collapse~\cite{chen2024neural}. Although correct pseudo-labels provide benefits, the performance gains they offer at the immature stage of the model often fail to offset the detrimental impact of erroneous pseudo-labels.

\begin{proof}
In semi-supervised pseudo-label learning, the model parameters are updated by the gradient of the pseudo-label samples. Let the model parameters be $\theta$ and the loss function for a single sample be $\ell(f_\theta(x), y)$. Define the gradient of the correct and incorrect pseudo-labels as: 

\begin{equation}
g_c(x_i) = \nabla_\theta \ell(f_\theta(x_i), y_i)
\end{equation}
\begin{equation}
    g_w(x_i) = \nabla_\theta \ell(f_\theta(x_i), \hat{y}_i),\ \hat{y}_i \neq y_i
\end{equation}

The parameter update formula is:

\begin{equation}
\theta' = \theta - \eta \cdot g(x_i),
\end{equation}

Where $\eta$ is the learning rate and $g(x_i)$ is the gradient of the corresponding sample.  

The core difference lies in the gradient direction:

\begin{itemize}
    \item For the correct pseudo-label, $g_c$ is consistent with the true gradient direction and has a smaller magnitude when converging.
    \item For incorrect pseudo-labels, $g_w$ usually deviates from the true gradient direction and may even point in the direction of increasing risk, thereby pushing the model away from the optimal point.
\end{itemize}

The expected risk of the model can be expressed as:
\begin{equation}
\mathcal{R}(\theta) = \mathbb{E}_{(x,y)\sim \mathcal{D}} \big[\ell(f_\theta(x), y)\big].
\end{equation}

After a small parameter update, the first-order Taylor expansion of the risk is:

\begin{equation}
\mathcal{R}(\theta - \eta g) 
\approx \mathcal{R}(\theta) - \eta \nabla_\theta \mathcal{R}(\theta)^\top g
+ \frac{1}{2} \eta^2 g^\top H_\mathcal{R} g,
\end{equation}
where $H_\mathcal{R}$ is the Hessian matrix of expected risk.

When the model is close to convergence, $\nabla_\theta \mathcal{R}(\theta) \approx 0$ and $\|g_c\|$ is extremely small, so:
\begin{equation}
\Delta \mathcal{R}_c \approx \frac{1}{2} \eta^2 g_c^\top H_\mathcal{R} g_c \approx 0,
\end{equation}
That is, correct pseudo labels have almost no negative impact on risk during the convergence phase.

Assume that the angle between $g_w$ and the true gradient is $\phi$. Upon convergence, $\|g_c\|\approx 0$, the first-order term almost disappears, and the risk change is mainly dominated by the second-order term:

\begin{equation}
\Delta \mathcal{R}_w 
\approx \frac{1}{2} \eta^2 g_w^\top H_\mathcal{R} g_w.
\end{equation}

Since the wrong pseudo-labels are often located in the uncertainty area of the model, their gradient amplitude is large, and under the local convexity assumption:

\begin{equation}
\Delta \mathcal{R}_w > 0,
\end{equation}

The conclusion shows that incorrect pseudo-labels can significantly increase the expected risk. Gradient-based analysis shows that a conservative pseudo-label screening strategy must be adopted in semi-supervised learning to reduce the impact of incorrect pseudo-labels.

\end{proof}

However, due to the overconfidence issue inherent in pseudo-labels, it is challenging to determine a single confidence threshold that reliably reflects the overall trustworthiness of multiple factor judgments for a given sample. To address this, we propose a Conservative Pseudo-Labeling Strategy that ranks samples based on the minimum confidence across multiple factors and selects only those with high minimum confidence for training, effectively mitigating the propagation of erroneous pseudo-labels.

Even when a single model exhibits overconfidence, the ranking strategy ensures that samples adopted in early stages predominantly come from distribution regions where the model performs best, making it harder for erroneous pseudo-labels to enter the top-ranked list and thus reducing error reinforcement during training. Overconfidence in a single model often occurs in regions where the model is undertrained, motivating the introduction of multi-factor judgments:

\begin{equation}
    s_i = \min_k p_i^{(k)}
\end{equation}

As long as one factor “questions” the sample, its ranking score will significantly decrease, thereby preventing its selection.Confidence-based ranking and top-k selection effectively mitigate the overconfidence problem in pseudo-labeling by naturally establishing a dynamic, distribution-adaptive confidence threshold. The multi-factor minimum confidence strategy further amplifies this suppression effect, as local overconfidence in one factor can be vetoed by others. Nevertheless, this approach cannot completely eliminate overconfidence, since samples may still be selected if all factors simultaneously exhibit excessive confidence.

\subsection{Multi-source Information from LMM}

Although our differentiated-factor approach strives to fully exploit existing information, the model remains prone to overconfidence and confirmation bias when labeled data is extremely scarce or node feature dimensions are limited. To address this, we incorporate large multimodal models as external information sources, generating auxiliary descriptions from each node’s original features or associated content (e.g., textual multimodal data). These auxiliary descriptions serve as weakly correlated or independent judgment factors, contributing to the comprehensive pseudo-label selection process.

For each node $v_i$ to be predicted, we first extract the original modal data related to it, such as associated text (such as document titles, social posts, etc.) or image modal content. Then, we use the multimodal large model $M_{LMM}$ to generate the semantic description vector $\mathbf{d}_i$ of the node:

\begin{equation}
    \mathbf{d}_i = \mathcal{M}_{\text{LMM}}(\text{context}_i)
\end{equation}

where $\mathbf{d}_i$ represents the auxiliary description vector of the node, which is used to express its potential semantic category in external knowledge. Next, we use a lightweight classification head $g_\phi(\cdot)$ to map the description vector to the category space:

\begin{equation}
    \mathbf{p}_i^{(d)} = \text{softmax}(g_\phi(\mathbf{d}_i))
\end{equation}
The category distribution $\mathbf{p}_i^{(d)}$ based on the descriptive information is obtained, which can be used as a weakly correlated auxiliary factor for pseudo-label selection.

Furthermore, a GCN can be used to combine the original information $x_i$, the graph structure neighborhood $\mathcal{N}(v_i)$ and the auxiliary description vector $\mathbf{d}_i$ and then map them to the category space:
\begin{equation}
    \mathbf{p}_i^{(cat)} = \text{softmax}(GCN(x_i,\mathcal{N}(v_i),\mathbf{d}_i))
\end{equation}
Obtain a more comprehensive weakly correlated auxiliary factor $\mathbf{p}_i^{(cat)}$.

While incorporating auxiliary factors generated by large multimodal models (LMMs) can enrich the signals for pseudo-label selection, their standalone prediction accuracy in node classification tasks often falls short of graph neural networks (GNNs) based on original data. Directly integrating these auxiliary factors as equally weighted judgment sources within the consistency selection mechanism leads to two outcomes:

\begin{itemize}
\item Slight improvement in confidence for correct samples: auxiliary factors may reinforce the reliability of pseudo-labels on nodes already assigned high confidence.
\item Amplification of erroneous disagreement: due to higher noise in auxiliary factor predictions, inconsistencies with the main model reduce the rate of right pseudo-label acceptance, degrading overall performance.
\end{itemize}

To reconcile this trade-off, we propose an Accountability Scoring Mechanism. ASM leverages auxiliary factors to enhance the ranking of trustworthy pseudo-labels while avoiding their direct interference in the consistency filtering, thereby balancing the two above.

First, based on the multi-factor consistency mechanism mentioned above, we filter out a preliminary pseudo-label set $S_u$ from the unlabeled dataset $D_u$. At this time, the judgment factors used for screening do not include the auxiliary factor prediction $\mathbf{p}_i^{(d)} \ or \ \ \mathbf{p}_i^{(cat)}$, avoiding low-precision information sources from interfering with the initial screening.For the nodes $v_i \in S_u$ that pass the initial screening, we use auxiliary factors to perform weighted accountability scoring on their pseudo-label confidences.

We design the accountability scoring function:
\begin{equation}
    \tilde{s}_i = s_i + \lambda \cdot \sum_k^K \mathbb{I}(\mathbf{p}_i^{(k)}=\hat{y_i} ) \cdot a_i^k
\end{equation}

Where $s_i$ is the confidence score of the primary factor. $\tilde{s}_i$ is the score after accountability scoring. $\lambda$ is the accountability coefficient, which balances the influence of auxiliary factors. $k \in K$ represents the number of different auxiliary factors. $\mathbb{I}(\cdot)$ is the indicator function, indicating that accountability scoring is only performed when the predicted categories of the primary and auxiliary factors are consistent. $a_i^k$ represents the confidence score of the $k$th auxiliary factor for node $v_i$.

After obtaining the accountability scores $\tilde{s}_i$, reorder the initial pseudo-label set $S_u$ and select the top $\tau \%$ nodes as the final pseudo-label set $S_p$:

\begin{equation}
    \mathcal{S}_p = \text{Top}_\tau \big( \{ (v_i, \tilde{s}_i) \mid v_i \in \mathcal{S}_u \} \big)
\end{equation}

The final training loss is still weighted by the supervision loss and the pseudo label loss:
\begin{equation}
    \mathcal{L}_{\text{total}} = \mathcal{L}_{\text{sup}}(\mathcal{D}_l) 
+ \lambda_p \cdot \mathcal{L}_{\text{pseudo}}(\mathcal{S}_p)
\end{equation}

\section{Experiments}
\label{sec:Experiment}

We evaluate our method against various state-of-the-art GNN models on benchmark graph datasets and conduct a series of experiments to validate its superiority. The factor exhibiting the highest consistency is used as the output within the DiFac framework for evaluation. All experiments in this section utilize GPT-4o as the large multimodal model (LMM).

\subsection{Datasets and Experimental Settings}

We evaluate the performance of our proposed DiFac on multiple benchmark datasets including Cora, Citeseer, PubMed, and portions of OGBN-Arxiv~\cite{sen2008collective}. These citation networks represent documents as nodes and citation relationships as edges, with node features given by bag-of-words vectors corresponding to academic domains. Additionally, we retrieved the article titles and abstracts for each node in these datasets, which convert to auxiliary descriptive vectors generated via large multimodal models (LMMs). Following standard semi-supervised learning protocols, we adopt the conventional train, validation, and test splits for each dataset.Table~\ref{tab:statistics_data} gives statistics summarizing the four datasets.

\begin{table}[http]
    \centering
    \caption{Dataset Statistics.}
    \begin{tabular}{c c c c c}
    \toprule
    Dataset & $n$  & $m$ & $c$ & Standard Segmentation\\
    \midrule
    Cora & 2,708 & 1,433 & 7 & 140/500/1,000  \\
    Citeseer & 3,327 & 3,703 & 6 & 120/500/1,000  \\
    Pubmed & 19,717 & 500 & 3 & 60/500/1,000   \\
    OGBN-Arxiv & 2,585 &128 & 8 & 160/500/1,000 \\
    \bottomrule
    \end{tabular}

    \label{tab:statistics_data}
\end{table}

The neural architecture of DiFac, referred to as the backbone, offers considerable flexibility. It can incorporate various mainstream graph neural networks as the backbone, such as GCN~\cite{kipf2016semi}, GAT~\cite{velivckovic2017graph}, GraphSAGE~\cite{hamilton2017inductive}, GCNII~\cite{chen2020simple}, and BiGCN~\cite{wang2024}. We will present experiments with each of these architectures in subsequent sections. For convenience, the default backbone network is set to GCN.


\subsection{Direct Effectivity of information sources}

First, we observe the performance of direct training based on each information source and its combination. Each dataset in this paper shows two information sources: the original attribute information $x_i$ and structural information $\mathcal{N}(v_i)$ of the citation network, and the auxiliary description information $\mathbf{d}_i$ generated by the LMM based on the article title and abstract.

We selected GCN~\cite{kipf2016semi}, GAT ~\cite{velivckovic2017graph}, GCNII ~\cite{chen2020simple}, GraphSAGE~\cite{hamilton2017inductive}, and BiGCN~\cite{wang2024} as the backbone networks. The results are shown in Table \ref{tab:exp_LLMdata}. Bold data indicate the best accuracy.

Where $(x_i,\mathcal{N}(v_i))$ represents the original graph attributes and structure input, $(x_i,\mathcal{N}(v_i),\mathbf{d}_i))$ adds auxiliary description information to the original input, $(\mathbf{d}_i),\mathcal{N}(v_i))$ contains both graph structure information and auxiliary description information, while $(\mathbf{d}_i)$ only contains auxiliary description information.

\begin{table*}
    \centering
    \caption{Augmented Data Effectiveness Experiments $(\%)$.}

	\begin{tabular*}{0.8\textwidth}{@{\extracolsep{\fill}} cc|cccc } \toprule
       Input & Methods & Cora  & Citeseer & Pubmed & Ogbn-arxiv \\
    \midrule
 \multirow{5}[0]{*}{$(x_i,\mathcal{N}(v_i))$} 
 &  GCN~\cite{kipf2016semi}  & 81.96 & \boxed{\textbf{64.98}} & \boxed{\textbf{75.65}} & 82.69 \\
 &   GAT~\cite{velivckovic2017graph}  & 81.24 & 61.22 & 72.47 & \boxed{84.15} \\
 &   GCNII~\cite{chen2020simple}  & \boxed{\textbf{83.97}}  & 62.97 & 74.79 & 79.26 \\
 &   SAGE~\cite{hamilton2017inductive} & 77.61 & 61.32 & 72.19 & 81.04 \\
 &   BiGCN~\cite{wang2024} & 81.15 & 57.16 & 73.59 & 76.04 \\
\midrule
 \multirow{5}[0]{*}{$(x_i,\mathcal{N}(v_i),\mathbf{d}_i)$} &    
 GCN & 82.11 & 60.64 & 70.12 & 84.94 \\
 &   GAT & 81.38 & 60.48 & 70.88 & \boxed{\textbf{87.67}} \\
&    GCNII & \boxed{83.75} & 62.38 & 75.26 & 85.82 \\
  &  SAGE  & 81.04 & 59.55 & 67.02 & 84.96 \\
 &   BiGCN & 82.98 & \boxed{66.02} & \boxed{74.97} & 81.16 \\
\midrule
    \multirow{5}[0]{*}{$(\mathcal{N}(v_i),\mathbf{d}_i))$} &   GCN & 82.01 & 59.48 & 66.85 & 81.98 \\
&    GAT & 81.25 & 58.53 & 68.83 & \boxed{84.4} \\
&    GCNII & \boxed{83.67} & 60.32 & \boxed{71.07} & 82.54 \\
&    SAGE & 80.08 & 58.47 & 63.56 & 82.38 \\
&    BiGCN & 81.57 & \boxed{61.12} & 67.58 & 78.02 \\
 \midrule
 $(\mathbf{d}_i)$ & $MLP$ & 69.41 & 60.92 & 53.23 & 72.35 \\
        \bottomrule
    \end{tabular*}
    \begin{tablenotes}
    \item  $\boxed{\quad }$:  The maximum classification accuracy within the same Input.
   \end{tablenotes}
    \label{tab:exp_LLMdata}
\end{table*}

\subsection{Standard Framework Performance}

In this section, we conduct experiments with DiFac using three differentiated factors and two auxiliary judgment factors, demonstrating its performance across different backbone networks and datasets. The results are summarized in Table~\ref{tab:exp_standard}. On the Cora dataset, DiFac employing GCNII as the backbone achieves the highest accuracy of 85.86$\%$, representing an improvement of 1.89$\%$ over the standalone backbone network. Meanwhile, on the Citeseer dataset, using GCN as the backbone yields a 6.14$\%$ accuracy gain. In the remaining two datasets, the improvements achieved by DiFac are less pronounced, which appears to be related to the feature dimensionality of nodes in these datasets. We will further analyze this issue in subsequent experiments.

\begin{table*}
    \centering
    \caption{Standard Framework Experiment ($\%$).}
   
	\begin{tabular*}{0.9\textwidth}{@{\extracolsep{\fill}}l c c c c c} \toprule
        Framework & Backbone & Cora & Citeseer & Pubmed & Ogbn-arxiv \\
        \midrule
        \multirow{5}[0]{*}{DiFac}
        & GCN & 85.81  & \textbf{71.12} & 78.21      & 84.59 \\
        & GAT & 84.71  & 70.26        & 77.63        & \textbf{85.96}\\
        & GCNII & \textbf{85.86} & 69.87 & \textbf{78.43} & 84.16 \\
        & SAGE & 84.46  & 69.46 & 77.18    & 83.46 \\
        & BiGCN & 83.75 & 70.67 & 76.80   & 82.26 \\
        \bottomrule
    \end{tabular*}
    \label{tab:exp_standard}
\end{table*}

\subsection{Comparison against SSL}
In this section, our DiFac framework adopts the same configuration as in the previous experiments, using GCN as the backbone network.

To benchmark against state-of-the-art graph-based semi-supervised learning (SSL) methods, we compare DiFac with six advanced approaches. Self-training and co-training~\cite{li2018deeper} propose different pseudo-label selection strategies; here, we use the classical self-training method as well as the intersection of pseudo-label sets from co-training and self-training. M3S~\cite{sun2020multi} employs a self-checking mechanism combining deep clustering (deepcluster~\cite{caron2018deep}) and alignment to select pseudo-labels. InfoGNN~\cite{li2023informative} introduces an information-theoretic measure to select high-quality pseudo-labels. $CG^3$\cite{wan2021contrastive-1} contrasts node representations from GNN and generative models. CGPN\cite{wan2021contrastive-2} utilizes contrastive Poisson networks to enhance GNN learning with limited labeled nodes. GCN-PCL~\cite{lu2025pseudo} leverages contrastive learning to guide the model in identifying whether pairs of nodes belong to the same class, thereby improving pseudo-label reliability.

As shown in Table~\ref{tab:exp_Performence}, our DiFac achieves significant gains on Cora and Citeseer datasets, improving accuracy by 3.85$\%$ and 6.14$\%$ respectively. On Pubmed and OGBN-Arxiv, DiFac performs competitively but does not surpass other advanced methods. We attribute this to the feature dimensionality: nodes in Cora and Citeseer have 1433 and 3703 dimensions respectively, whereas Pubmed and OGBN-Arxiv have only 500 and 128 dimensions. Since DiFac aims to extract multiple differentiating judgment factors from the same data source, insufficient intrinsic diversity in node features may cause forced factor extraction to degrade to a single model or even worsen performance.

\begin{table*}
    \centering
    \caption{Comparison against SSL ($\%$).}
	\begin{tabular*}{\textwidth}{@{\extracolsep{\fill}} ccccc } \toprule
         Methods & Cora  & Citeseer & Pubmed & Ogbn-arxiv \\
        \midrule
    Intersection & 79.72(2.23$\downarrow$) & 65.77(0.78$\uparrow$) & 75.15(0.50$\downarrow$) & 84.21(0.06$\uparrow$) \\
    Self-training & 80.25(1.71$\downarrow$) & 63.83(1.15$\downarrow$) & 74.97(0.68$\downarrow$) & 84.48(0.23$\uparrow$) \\
    InfoGNN & 82.94(0.98$\uparrow$) & 69.44(4.70$\uparrow$) & 76.31(0.67$\uparrow$) & 84.79(0.64$\uparrow$)\\
     $CG^3$ & 83.47(1.78$\uparrow$) & 69.68(4.94$\uparrow$) & 78.28(2.63$\uparrow$) & 85.27(1.11$\uparrow$)\\
      CGPN & 83.26(1.40$\uparrow$) & 68.73(3.75$\uparrow$) & 77.88(2.23$\uparrow$) & 85.06(0.90$\uparrow$)\\
    GCN-PCL & 84.36(2.40$\uparrow$) & 69.63(4.65$\uparrow$) & \textbf{80.37}(\underline{4.71$\uparrow$}) & \textbf{85.49}(\underline{1.34$\uparrow$})\\
       \midrule
      DiFac & \textbf{85.81}(\underline{3.85$\uparrow$}) & \textbf{71.12}(\underline{6.14$\uparrow$}) & 78.21(2.56$\uparrow$) & 84.59(0.44$\uparrow$)\\
        \bottomrule
    \end{tabular*}
   
    \begin{tablenotes}
    \item  $\uparrow$:  Increment of classification accuracy for the corresponding backbone model.
    \item  $\underline{\quad}$:  The maximum increment.
   \end{tablenotes}
    \label{tab:exp_Performence}
\end{table*}

\begin{figure*}[htbp]
\centering
\subfigure[Cora]
{\label{fig:label_rate_cora}
\includegraphics[width=0.98\textwidth]{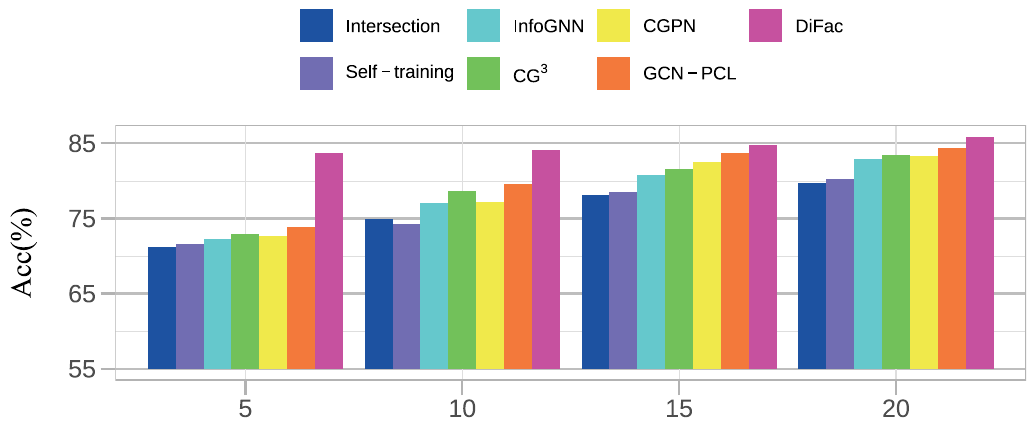} }
\subfigure[Citeseer]
{\label{fig:label_rate_citeseern}
\includegraphics[width=0.98\textwidth]{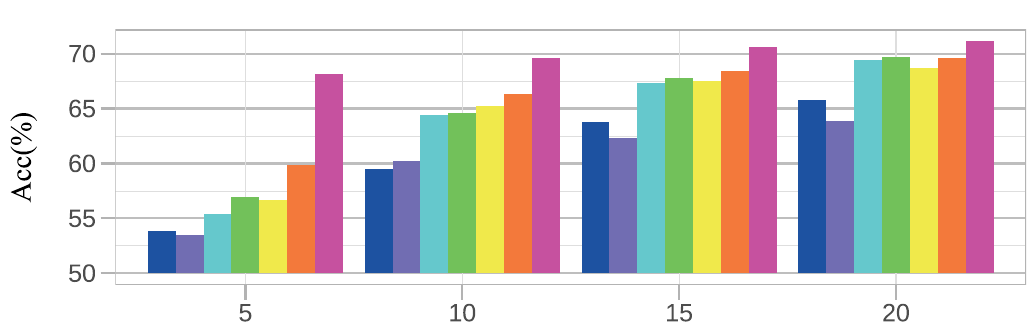} }
\subfigure[Pubmed]
{\label{fig:label_rate_pubmed}
\includegraphics[width=0.98\textwidth]{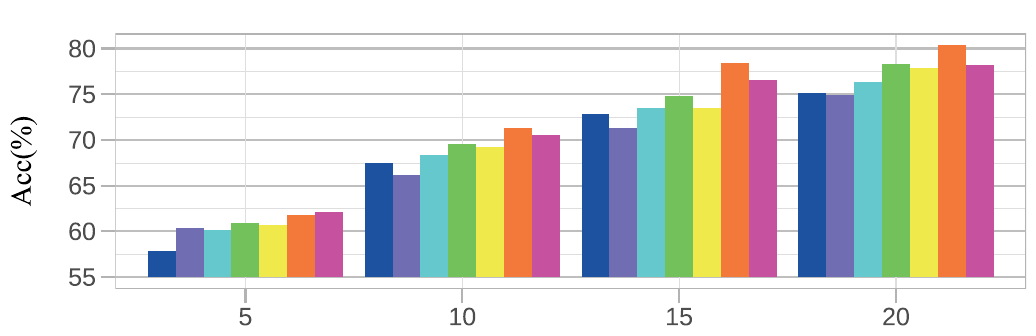} }
\caption{Label Rate Experiments $(\%)$.} 
\label{fig:exp_label_rate}
\end{figure*}

In the standard training set splits, each class contains 20 labeled samples. To evaluate performance under extremely low label rates, we conduct experiments with training samples per class reduced to $\{5, 10, 15, 20\}$. The results are presented in Figure~\ref{fig:exp_label_rate}. It can be observed that when only 5 training samples per class are available, the accuracy of nearly all models drops sharply; however, DiFac maintains a robust accuracy of 83.12$\%$ on the Cora dataset. This resilience stems from the stability provided by multi-source information. While reduced label rates degrade the representation capacity of original data features, the auxiliary descriptive information generated by large multimodal models (LMMs) remains largely unaffected.

\subsection{Differentiation Factors Study}

\begin{figure*}[htbp]
\centering
{\includegraphics[width=0.98\textwidth]{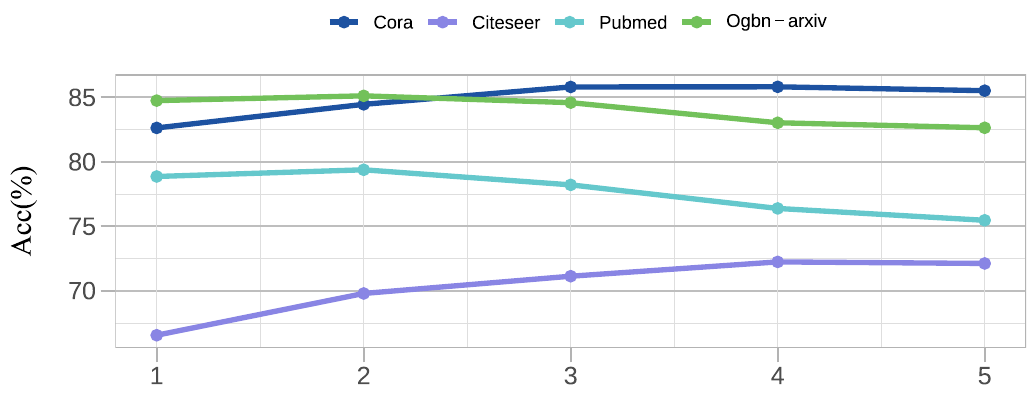} }
\caption{Differentiation Factors Experiments $(\%)$.} 
\label{fig:exp_d_factors}
\end{figure*}

In this experiment, we investigate the efficiency of mining different numbers of differentiated factors across four datasets, using GCN as the backbone network. The results are presented in Figure~\ref{fig:exp_d_factors}. On Citeseer, the model exhibits a pronounced performance improvement as the number of differentiation factors increases. Similarly, Cora also demonstrates an upward trend. In contrast, Pubmed and Ogbn-arxiv display a clear performance decline. These observations further support our earlier assertion that the effectiveness of mining differentiation factors depends on the intrinsic diversity of node features.

\subsection{Ablation Study}

To assess the contribution of each component in the framework, we conduct experiments comparing different differentiation construction methods, examining various confidence ranking strategies, and evaluating the impact of the top-ratio selection, the number of pseudo-labeling iterations, and the incorporation of auxiliary factors.

\subsubsection{Differentiated Construction Study}

We compare three differentiated imputs construction methods in terms of their impact on accuracy: (1) introducing differentiation markers into node attributes, (2) randomly inverting the values of certain attribute dimensions, and (3) randomly exchange the values of certain attribute dimensions, GCN is adopted as the backbone network, and the results are summarized in Table~\ref{tab:differ_construction}. Although the differences among the three methods are not pronounced, the differentiation-marker approach consistently outperforms the other two across all three datasets. This suggests that minimizing perturbations to the original data leads to better performance.

\begin{table}
    \centering
    \caption{Differentiated Construction Experiment($\%$).}
    \begin{threeparttable}
	\begin{tabular*}{0.47\textwidth}{@{\extracolsep{\fill}}l c c c} \toprule
        Diff-Methods & Cora & Citeseer &Pubmed \\
        \midrule
       differentiation marker & \textbf{85.81} & \textbf{71.12} & \textbf{78.21} \\
       random reverse & 84.74 & 70.71 & 77.05 \\
        random exchange & 85.32 & 70.89 & 77.75 \\
        \bottomrule
    \end{tabular*}
    \end{threeparttable}
\label{tab:differ_construction}
\end{table}

\subsubsection{Rank Strategy Study}

In this experiment, we employ GCN as the backbone network to examine the effects of different ranking strategies and varying initial top-ratios on model performance, as well as the influence of incorporating auxiliary factors. Here, the top-ratio refers to the initial selection ratio, which eventually reaches 90$\%$ through iterative updates. The results, shown in Figure~\ref{fig:exp_rank_line}, clearly indicate that the minimum-confidence ranking strategy offers a notable advantage, mitigating the unreliability of confidence scores caused by model overconfidence. Furthermore, the inclusion of auxiliary factors always yields performance gains.

\textbf{\begin{figure*}[htbp]
\centering
\subfigure[With Cofactors]
{\label{fig:exp_rank_line_asis}
\includegraphics[width=0.48\textwidth]{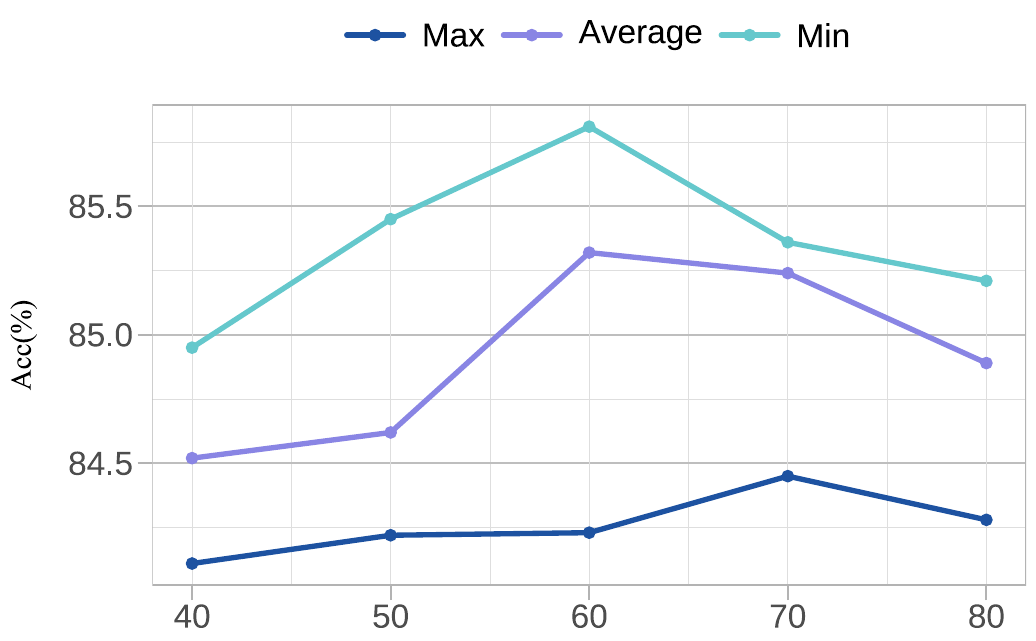} }
\subfigure[Without Cofactors]
{\label{fig:exp_rank_line_noasis}
\includegraphics[width=0.48\textwidth]{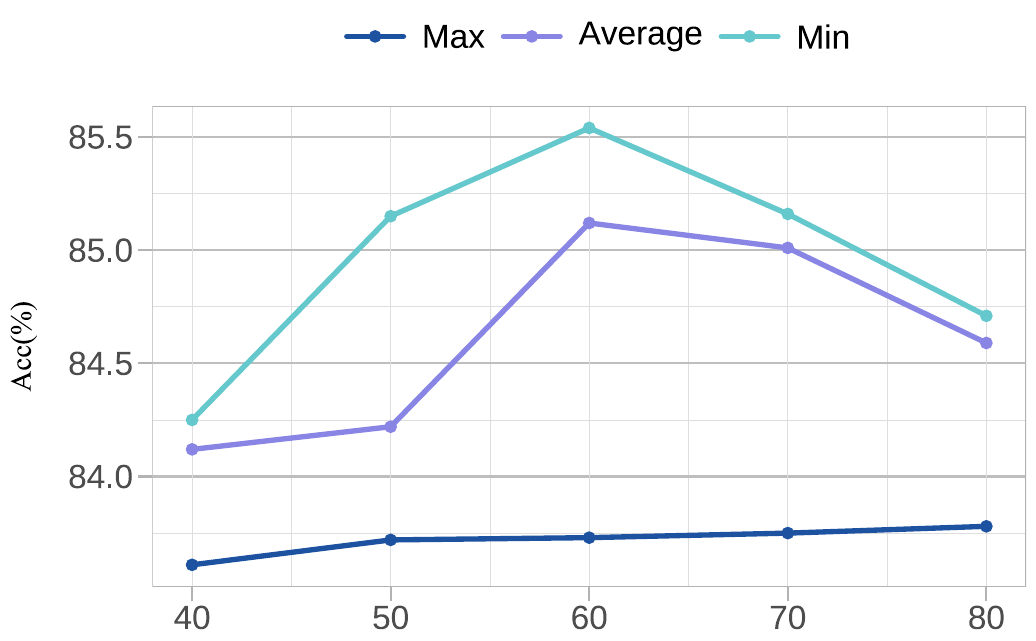} }
\caption{Rank Strategy \& Top rate Experiments.} 
\label{fig:exp_rank_line}
\end{figure*}}

We further observe the relationship between the number of iterations and the ranking ratio. The experimental results are shown in Figure \ref{fig:exp_rank_hot}.

\textbf{\begin{figure*}[htbp]
\centering
\subfigure[With Cofactors]
{\label{fig:exp_rank_hot_asis}
\includegraphics[width=0.40\textwidth]{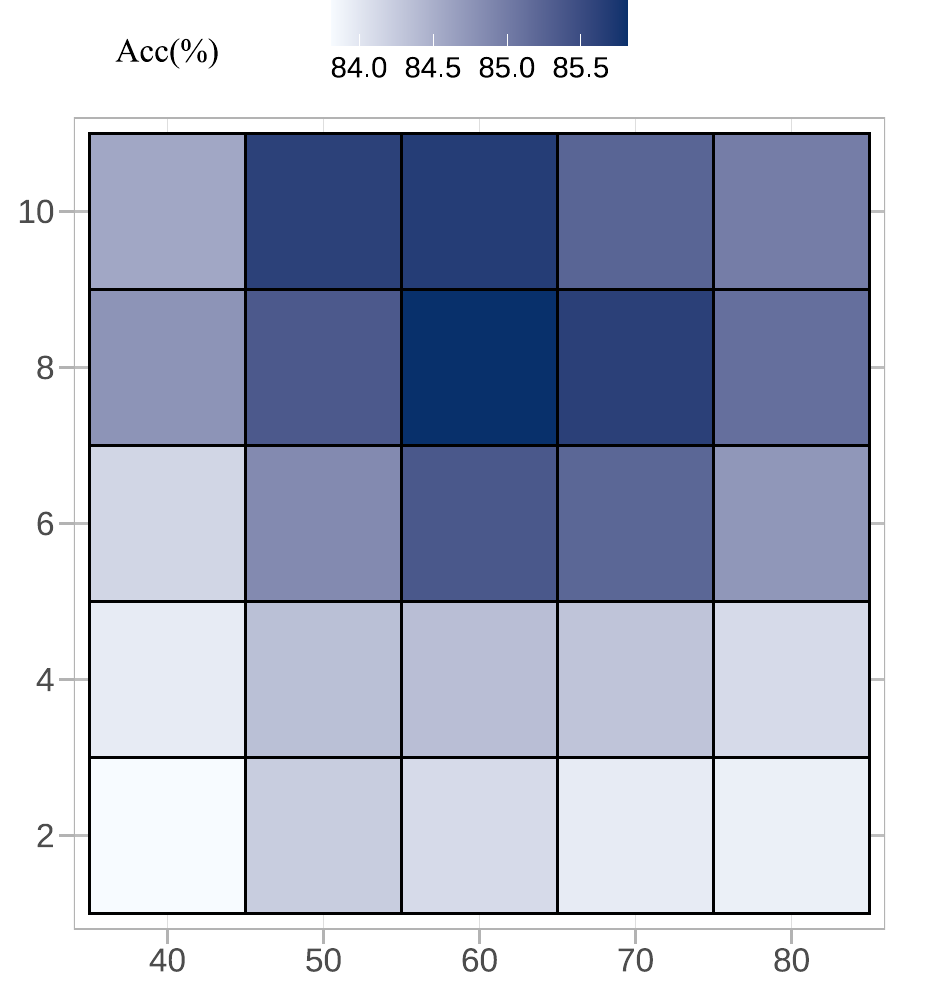} }
\subfigure[Without Cofactors]
{\label{fig:exp_rank_hot_noasis}
\includegraphics[width=0.40\textwidth]{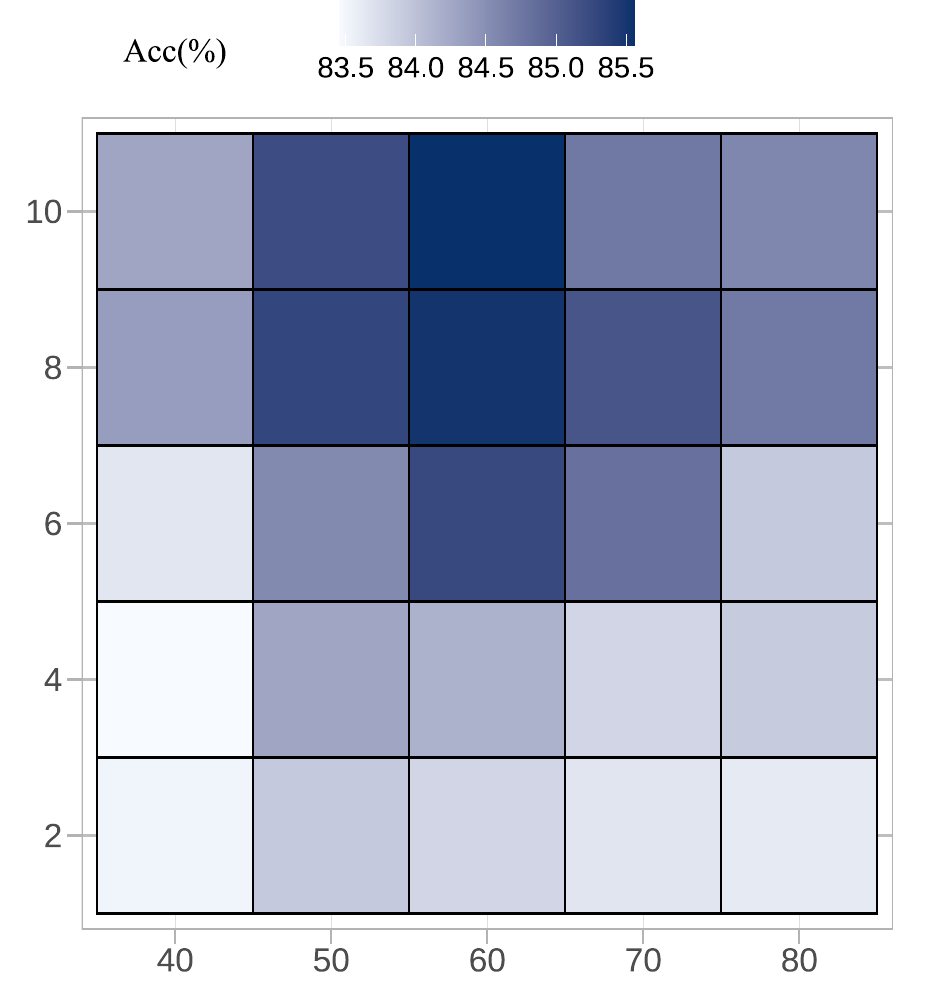} }
\caption{Number of iterations \& Top rate Experiments.} 
\label{fig:exp_rank_hot}
\end{figure*}}

\subsection{Conceit Evaluation Study}
We compare against the Intersection model to evaluate the reliability of the adopted pseudo-labels, measured as the accuracy of the selected pseudo-labels (90$\%$) on the test set. GCN is used as the backbone network, and the results are presented in Table~\ref{tab:conf_show}.

\begin{table}
    \centering
    \caption{Confidence Consistency Experiment($\%$).}
    \begin{threeparttable}
	\begin{tabular*}{0.47\textwidth}{@{\extracolsep{\fill}}l c c c} \toprule
        Methods & Cora & Citeseer &Pubmed \\
        \midrule
        DiFac & 85.81 & 71.12 & 78.21 \\
       Intersection & 79.72 & 65.77 & 75.15 \\
        DiFac$_{conf}$ & 86.74 & 73.45 & 79.85 \\
       Intersection$_{conf}$ & 81.84 & 66.71 & 77.05 \\
        \bottomrule
    \end{tabular*}
    \end{threeparttable}
\label{tab:conf_show}
\end{table}

We define an conceit metric as the difference between (i) the cosine similarity of an overconfident misclassified sample to the mean representation of all correctly classified samples in the predicted class and (ii) its cosine similarity to the mean representation of all correctly classified samples in its true class. A larger value of this metric indicates a higher degree of overconfidence:
\begin{equation}
\begin{aligned}
     &Conceit =  \\
    & \sum_{(y_i=c \ ,\ \hat{y}_i=k)} \big(sim(z_i\ ,\sum_{\hat{y}_j=y_j=c}z_j) - sim(z_i \ ,\sum_{\hat{y}_j=y_j=k}z_j)\big)
\end{aligned}
\end{equation}

Where $c\not=k$, $z_i$ represents the representation vector of node $v_i$, and $sim(\cdot)$ means calculating the cosine value of two vectors.

Using GCN as the backbone network, the experimental results are presented in Table~\ref{tab:conf_show}. We observe that DiFac exhibits a lower degree of conceit on the Cora and Citeseer datasets, but a higher degree on Pubmed. This indicates that, on the former two datasets, DiFac is more effective in discerning the validity of pseudo-labels.

\begin{table}
    \centering
    \caption{Conceit Evaluation Experiment($Conceit$).}
    \begin{threeparttable}
	\begin{tabular*}{0.47\textwidth}{@{\extracolsep{\fill}}l c c c} \toprule
        Methods & Cora & Citeseer &Pubmed \\
        \midrule
        DiFac & 0.13 & 0.08 & 0.34 \\
       Intersection & 0.25 & 0.21 & 0.27 \\
        \bottomrule
    \end{tabular*}
    \end{threeparttable}
\label{tab:conceit_show}
\end{table}

\subsection{Random Mask Experiments}
This experiment evaluates the effectiveness of DiFac under extreme noise conditions by randomly masking a certain proportion of node attribute dimensions, with masking ratios set to {10, 30, 50, 70, 90}. Using GCN as the backbone network, the results are reported in Table~\ref{tab:mask_filter}. Even with 90$\%$ of the attribute dimensions masked, DiFac still achieves an accuracy of $72.93$, demonstrating strong robustness.

\begin{table*}
     \centering
     \caption{Random Mask Experiments ($\%$).}
     \resizebox{\textwidth}{!}{ %
	\begin{tabular*}{\textwidth}{@{\extracolsep{\fill}} c|ccccc } \toprule
        Datasets & 10    & 30    & 50    & 70    & 90  \\ 
    \midrule
    Cora
          &  84.24 & 82.14 & 80.27 & 78.24 & 72.93 \\
    \midrule
    Citseer
          & 69.81 & 62.41 & 58.14 & 57.09 & 52.87 \\
    \midrule
    Pubmed
          &  74.45 & 73.76 & 69.37 & 61.32 & 51.02 \\
    \midrule
    OGBN-arxiv
          &  76.81 & 72.41 & 68.14 & 62.09 & 57.47 \\
        \bottomrule
    \end{tabular*}}
 
\label{tab:mask_filter}
\end{table*}

\subsection{Complexity Analysis}
\label{subsec:complexity_analysis}

In this study, our model employs a single neural network architecture to extract multiple differentiated judgment factors, thereby avoiding the computational overhead associated with parallel inference of multiple sub-models in traditional ensemble methods. Specifically, if each sub-model has an inference complexity of $O(f(n))$, the overall complexity of an ensemble with kkk models is $O(k \cdot f(n))$. In contrast, our approach integrates multi-factor modeling within a unified network through internal modular partitioning or parameter sharing, maintaining an inference complexity of $O(f(n))$ and significantly reducing resource consumption during inference.
Furthermore, during training, we incorporate auxiliary descriptive vectors generated by multimodal large language models to enhance the expressiveness of differentiating factors. The time complexity of this component mainly depends on the inference efficiency of the large model itself, denoted as $O(g(n))$, where $g(n)$ represents the processing complexity of the multimodal model for input samples. Since this step is typically performed offline and auxiliary vectors can be stored and reused, its impact on overall training time remains controllable.

\section{Conclusion}
\label{sec:Conclusion}  

In this work, we propose a method for mining latent decision factors from implicit information by constructing identical data samples into different classes, thereby guiding a graph neural network (GNN) to learn differentiated decision factors from single-source data. To better exploit these factors, we introduce a ranking strategy and an auxiliary factor scoring strategy, which, when combined, can effectively enhance graph-based SSL performance. Extensive experiments validate the effectiveness of our approach, while comprehensive ablation studies elucidate the importance of each design choice. Notably, our method is particularly well-suited for datasets with higher feature dimensionality or greater intrinsic diversity. In the future, we aim to further improve and extend our work, with the goal of applying it to the domain of automatic data labeling.

\bibliographystyle{IEEEtran}
\bibliography{DiFac}

\end{document}